{\noindent\ignorespaces\color{#1}\textbf{#2:}}%
{\par\noindent\ignorespacesafterend}

\newcommand{\eg}{\emph{e.g.,}}
\newcommand{\ie}{\emph{i.e.,}}

\newcommand{\etal}{\emph{et. al}}
\documentclass[usenames,dvipsnames,conference]{IEEEtran}  

\usepackage{times}
\usepackage{epsfig}
\usepackage{graphicx}
\usepackage{amsmath}
\usepackage{amssymb}

\usepackage{gensymb}
\usepackage{balance}
\usepackage{algorithm}
\usepackage[noend]{algpseudocode}
\usepackage{placeins}

\usepackage{siunitx}
\usepackage{tabu}
\usepackage{tabularx}
\usepackage{tablefootnote}
\usepackage{threeparttable}
\usepackage{mathtools}
\usepackage{epstopdf}
\usepackage{float}
\usepackage[T1]{fontenc}
\usepackage{enumitem}
\usepackage[percent]{overpic} 

\graphicspath{ {plots/} {diagrams/} {figures/}}

\usepackage[pagebackref=true,breaklinks=true,letterpaper=true,colorlinks,bookmarks=false]{hyperref}

\begin{document}


\title{ClickBAIT-v2: Training an Object Detector in Real-Time}

\author{\IEEEauthorblockN{Ervin Teng, Rui Huang, and Bob Iannucci}
\IEEEauthorblockA{Department of Electrical and Computer Engineering \\
Carnegie Mellon University\\
NASA Ames Research Park, Building 23 (MS 23-11), Moffett Field, CA 94035\\
{\tt \{ervin.teng, rui.huang, bob\}@sv.cmu.edu}}}

\maketitle
\begin{abstract} 

Modern deep convolutional neural networks (CNNs) for image classification and object detection are often trained offline on large static datasets. Some applications, however, will require training in real-time on live video streams with a human-in-the-loop. We refer to this class of problem as time-ordered online training (ToOT). These problems will require a consideration of not only the quantity of incoming training data, but the human effort required to annotate and use it. We demonstrate and evaluate a system tailored to training an object detector on a live video stream with minimal input from a human operator. We show that we can obtain bounding box annotation from weakly-supervised single-point clicks through interactive segmentation. Furthermore, by exploiting the time-ordered nature of the video stream through object tracking, we can increase the average training benefit of human interactions by 3-4 times.

\end{abstract}




\section{Introduction}
\label{sec:intro}
Today's workflow for creating a convolutional neural network (CNN) application, such as an object detector, is to (a) obtain training dataset(s), perhaps from the Web or gathered in the field; (b) upload these to a training server; (c) run the training algorithm; (d) obtain resulting weights from the training server and (e) deploy to the end device. This static, offline approach, while effective, is a limiting factor in many applications. Consider, for instance, CNNs onboard small autonomous systems, such as unmanned aerial systems (UAS). CNNs can dramatically reduce the mental burden of human operators by automating much of the video processing. However, statically-trained CNNs will leave the UAS ineffective and vulnerable when it faces conditions and changes in the environment unexpected during the model creation process. 

Imagine a UAS is trained to find and track a particular person during a search-and-rescue or surveillance operation. Perhaps it is discovered early into the mission that the CNN model deployed to the UAS is performing poorly at human detection from an overhead angle. Or, perhaps during the mission the target person changes clothes---and immediately, the UAS that was following him or her loses track. In a statically-trained model workflow, re-gathering imagery, re-training, and re-deploying the model is a cumbersome process, and would result in the termination of the mission. 

People, however, learn very differently. They are not only trained statically beforehand, but are also \textit{field-trained}, \ie{} they practice in situations resembling the usage scenario, with a trainer providing positive or negative reinforcement. Training does not terminate after the training phase---new information and experiences gathered ``on the job'' can also better the trainee. By analogy, we can imagine training a CNN-equipped UAS by showing it examples of what it's looking for (and what it's not looking for), over time building a better and better model---just like showing a child flash cards. Even during use, if the CNN behaves incorrectly, the user can correct its behavior. This type of training will require ingestion and annotation of data streams, such as video, in real-time. To do so, we must make special consideration of the human effort required to create image annotations. We call this class of problem \textit{time-ordered online training} (ToOT).




\begin{figure}[t] \centering
    \includegraphics[width=0.9\linewidth]{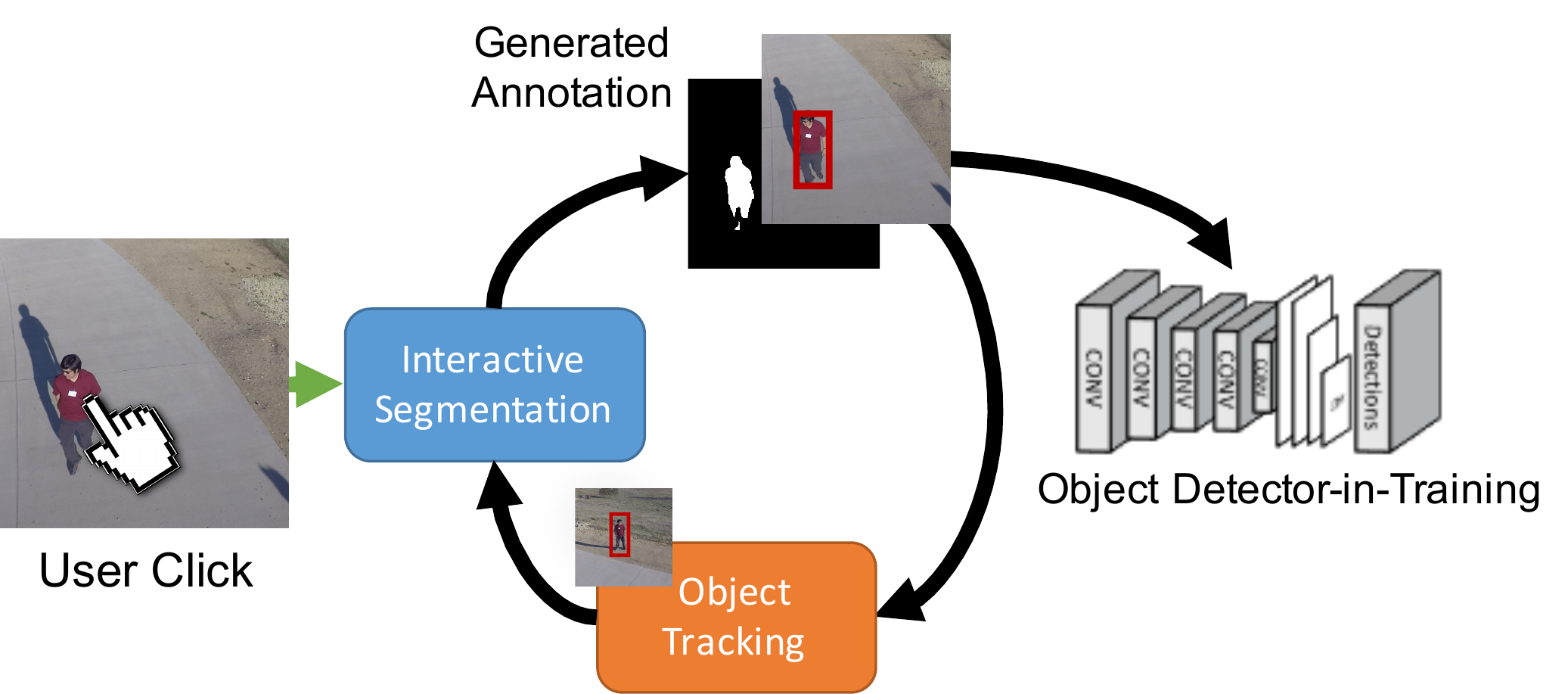}
    \caption{Overview of the ClickBAIT-v2 system. A user initiates an interaction by clicking on the desired target. Using the click and the image as input, an interactive segmenter segments out the object, which is used to derive a bounding box. This is used to train the detector for one round. It is also used to initialize an object tracker, which creates subsequent clicks without further input from the user.}
    \label{fig:clickbait_overview}
\end{figure}

In~\cite{Teng2017}, the authors present ClickBAIT, a system to perform online, real-time training of an image \textit{classifier} onboard a small UAS. In this paper, we present ClickBAIT-v2, extending the system to object detection without increasing the annotation difficulty. Figure~\ref{fig:clickbait_overview} shows how ClickBAIT-v2 assists and automates part of the annotation process so that time-ordered online training is feasible in real-time. The main contributions are:

\begin{itemize}[noitemsep]

\item A weakly-supervised method for using user clicks to obtain bounding box annotations, eliminating the need for drawing them on every image. 
\item A combined approach to greatly increase the effectiveness of human input during time-ordered online training, combining click-based supervision with object tracking. 
\end{itemize}

This paper is structured as follows. Section~\ref{sec:background} provides an overview of the state-of-the-art. Section~\ref{sec:architecture} describes the design of ClickBAIT-v2 and its subsystem. Section~\ref{sec:experiments_weakly} describes an evaluation of the weakly supervised bounding box generation, and Section~\ref{sec:experiments_clickbait} evaluates the effectiveness of ClickBAIT-v2 in reducing user interaction while training.



\section{Background and Prior Work}
\label{sec:background}


Training CNNs has traditionally been done offline, whether on large datasets such as ImageNet~\cite{Krizhevsky2012,Szegedy2014}, on synthetic datasets~\cite{Kiswani2016}, or on collected datasets~\cite{Guisti2016}. Active learning has recently been applied to human-in-the-loop deep image classification~\cite{Wang2017Cost-EffectiveClassification}, but requires the network to choose its images---something not possible with the continuous progress of time in a video sequence. Weakly-supervised training can train detection via class labels~\cite{Bilen16,Wang2014}, and unsupervised training of foreground detection can be achieved using non-learning methods~\cite{Croitoru2017UnsupervisedImages}, but these methods have an inherent class-level or instance-level ambiguity. We use user interaction via clicks to remove this ambiguity.

\textit{Fine-tuning} or \textit{transfer learning} is the process of taking pre-trained weights and refining them to fit a particular application, using the fact that many of the features within a trained network are common across applications~\cite{Yosinski2014}. In~\cite{Kading2016} investigate continuous learning as a sub-problem of fine-tuning, and conclude that continuous, online learning can be achieved by directly fine-tuning the network in small steps. We will examine our particular problem, time-ordered online training, as a subset of continuous learning.

Weakening the supervision, \eg from bounding boxes to clicks, required to train a model, is one way to make datasets easier to annotate. A prior approach to click supervision for object detection~\cite{Papadopoulos2017} does show significant success in reducing annotation time--however, their MIL-based approach is unsuited for online execution. Similarly, we found that multiple-interaction~\cite{Jain2016} and proposal-mining~\cite{Mettes2016} methods for weakening supervision unsuited for real-time use, opting for a single-click approach. 

\subsection{Measuring the Effectiveness of a User Interaction}

In~\cite{Teng2017}, the authors introduce time-ordered online training (ToOT) as the real-time training of a classifier on a live video stream, using human-in-the-loop annotations. They present a metric, \textit{training benefit}, that describes the effectiveness of each user interaction, defined in relation to the trained model's accuracy, with the premise that human annotation will quickly become the bottleneck during real-time training. 

In this paper, we extend this definition to include models-under-training other than classifiers, and define cumulative and incremental training benefit~\cite{Teng2017} in terms of a scalar performance value $P_i$. \textit{Performance} can indicate a number of metrics depending on the task at hand; for object detection, \textit{average precision} (AP) is a suitable metric. We reserve $P_0$ as the performance of the untrained network.


For a vector of $n$ video frames $V = (v_1, v_2, ..., v_n)$ and a corresponding vector of user interactions 
\begin{equation}
U = (u_1, u_2, ..., u_n),   u_i = \begin{cases} 
      1 & \text{user int. on frame $v_i$} \\
      0 & \text{otherwise}
   \end{cases}
\end{equation}
we define the \emph{cumulative training benefit} of frames up to and including frame $v_i$ as

\begin{equation}
\label{eqn:CTB}
CTB_{i} = \begin{dcases*} 
      P_i-P_0 & $i>0, i<=n$ \\
      \text{undefined} & otherwise
   \end{dcases*}
\end{equation}

Subsequently, the $CTB$ of the particular user interaction $u_i$ is given by

\begin{equation}
CTB_{u_i} = \begin{dcases*} 
      P_{k-1}-P_0 & $u_i = 1$ \\
      \text{undefined} & otherwise
   \end{dcases*}
\end{equation}
where $k$ is the index of the \textit{next} non-zero $u_k$, $k > i$. 

Likewise, the \emph{incremental training benefit} of a user interaction taken during training round $i$ can be defined by
\begin{equation}
ITB_{u_i} = \begin{dcases*} 
      P_{k-1}-P_{i-1} & $u_i = 1$ \\
      \text{undefined} & otherwise
   \end{dcases*}
\end{equation}

We can then compute the mean $ITB$ for training rounds $i...j$ as follows:

\begin{equation}
\label{eqn:ITB}
\overline{ITB}_{(i,j)} = \frac{\sum_{x=i}^{j} I_x}{\sum_{x=i}^{j} u_x}
\end{equation}

 \begin{equation}
I_{x} = \begin{dcases*} 
      ITB_{u_x} & $u_i = 1$ \\
      0 & otherwise
   \end{dcases*}
 \end{equation}
As in ~\cite{Teng2017ClickBAIT:Networks}, we make the stipulation that we terminate training once the model reaches some suitable final performance $P_f$. If we fix $P_f$, the mean $ITB$ for the sub-vector of user interaction $(u_1, ..., u_f)$ becomes a measure of the particular efficiency of a training algorithm, \textit{i.e.} how well each user interaction is being used.

\section{Training System Architecture}
\label{sec:architecture}
We consider the challenge of training a UAS to detect objects of interest within its field of view, while in-flight. While the system is operational, we perform online training for detection of a specific object instance. This object does not need to belong to a specific class, and the training process itself is class-agnostic. Furthermore, we make the constraint that any user interaction while the system is running must be a single click---and not time-consuming as dragging a bounding box. This section describes the three main components of  our system: the online-trained object detector; the segmenter, which allows bounding box supervision from a single click; and the object tracker, which multiplies each user interaction.

\subsection{Online Training of Object Detection}
\label{sec:ssd}
\textbf{Pre-training: }Due to hardware and real-time constraints, we chose the single shot object detector (SSD)~\cite{Liu2016} for its high framerate as compared to region proposal methods. We used the original SSD300 formulation and architecture with VGG-16, re-implemented in TensorFlow~\cite{BalancaSingleHttps://github.com/balancap/SSD-Tensorflow}. In order to convert the original 20-class model trained on PASCAL VOC2007+2012~\cite{EveringhamTheResults} to a 1-class model, we re-initialized and fine-tuned the final class prediction layers on the VOC2007+2012 trainval set with all objects given the same label. Because our targeted UAS use case often involves detecting small objects from altitude on the ground, we also randomly shrank, with probability 0.5, the images and bounding boxes to 1/4 of the dimensions, and placed them on a blank canvas as part of the data augmentation strategy. 
We conduct further data augmentation by applying random crop where the minimum intersection over union (IoU) of the cropped image to the bounding box is 0.25, mirror with probability 0.5 and perturbations in hue, contrast, saturation, and brightness. 

\textbf{Online Training: }We then began each online training sequence with the all-class fine-tuned model above. During this process, the model is further fine-tuned using the Adam~\cite{Kingma2015Adam:Optimization} optimizer, and learns instance-specific traits, including those outside of the original 20 classes. To imitate the shrinking image augmentation strategy described above during online training, we inserted into every batch one raw video frame, and one rescaled one. Since objects may be already very small in the raw frame, we rescaled the frames by factor $S$ in a dynamic way as given by Equation~\ref{eqn:dynamicscale}. 

\begin{equation}
\label{eqn:dynamicscale}
S = 10\frac{\max(h_{b}, w_{b})}{w_{i}}
\end{equation}
where $w_b$, $h_b$ is the width and height of the initial bounding box, and $w_i$ is the width of the (square) initial image. The image is then placed randomly on a canvas of dimensions $(h_i, w_i)$ of the mean values of the VOC training set. We then perform the same data augmentation steps as in pre-training to a duplicate pair of the normal and shrunken images, bringing each online batch to a total size of four. 

\subsection{Bounding Box Supervision from Clicks}
\label{sec:click_seg}

\begin{figure} \centering
    \includegraphics[width=0.9\linewidth]{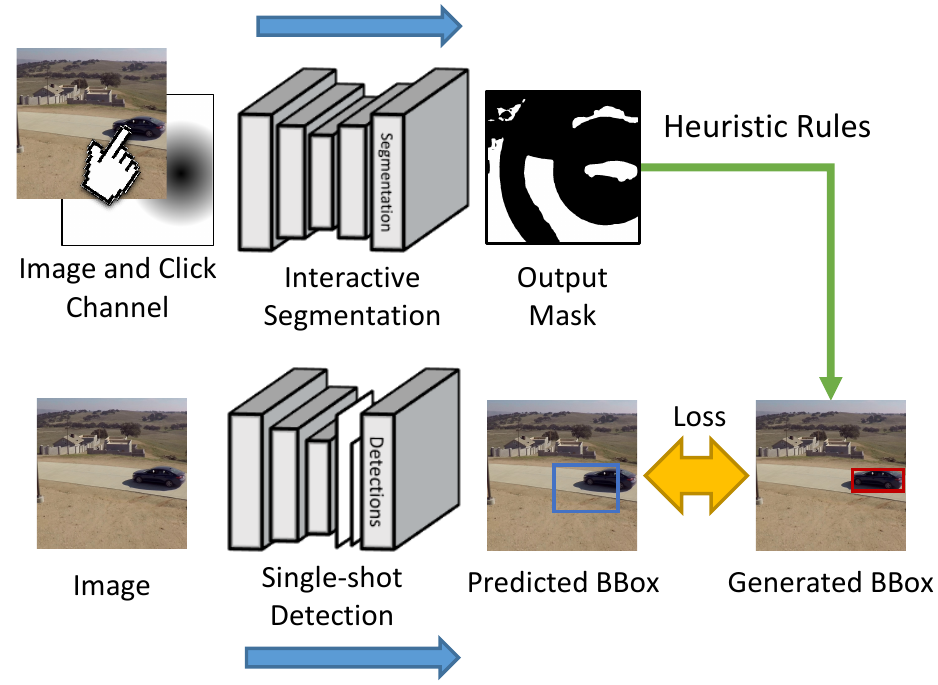}
    \caption{Diagram of interactive segmentation used to create bounding box annotations. Images are annotated with a single click, which is turned into a distance (Voronoi) channel~\cite{Xu2016} and appended to the image. The segmenter generates a mask from this input, from which a bounding box is created after filtering and heuristics. This bounding box is used to train the SSD network.}
    \label{fig:segment_click}
\end{figure}

Conducting ToOT for object detection would require drawing bounding boxes for every frame in the video stream in near real-time, which is not realistic or safe for a human operating a UAS. In order to alleviate this difficulty, we adopt a customized interactive segmentation technique to automatically generate a high-quality class-agnostic object bounding box annotation from a single user click, described in Figure~\ref{fig:segment_click}. Specifically, we modified the architecture from Xu \etal \cite{Xu2016} to only include positive clicks. In contrast to saliency detection, the segmenter produces a mask containing only the single object instance specified by the clicks, even if it is not the most salient within the image. With the binary mask output, we can derive a bounding box suitable for training detectors. We present the performance of the interactive segmentation and the weakly-supervised training of a detector in Section~\ref{sec:experiments_weakly}. 

During online training, we take three steps to reduce noise in the bounding box generated from the segmenter. First, the segmenter is provided input at full video resolution (720x720 pixels), rather than the input resolution of SSD300. Second, if the segmenter produces multiple regions, we select only the region within which the click landed, or the closest region to the click. The smallest bounding box that encompasses the region is created. If \textit{no} region is produced by the segmenter, a small bounding box of 20x20 pixels is created around the click. This last case is a very rare occurrence. Finally, the segmenter has issues with long, thin objects, often cutting off part of the object. Therefore, if the produced bounding box's smallest side is less than one tenth of the image dimension, we increase both dimensions of the bounding box by 20\%.

\subsection{Object Tracking for Training Assistance}
\label{sec:optical_flow}
\begin{figure} \centering
    \includegraphics[width=0.8\linewidth]{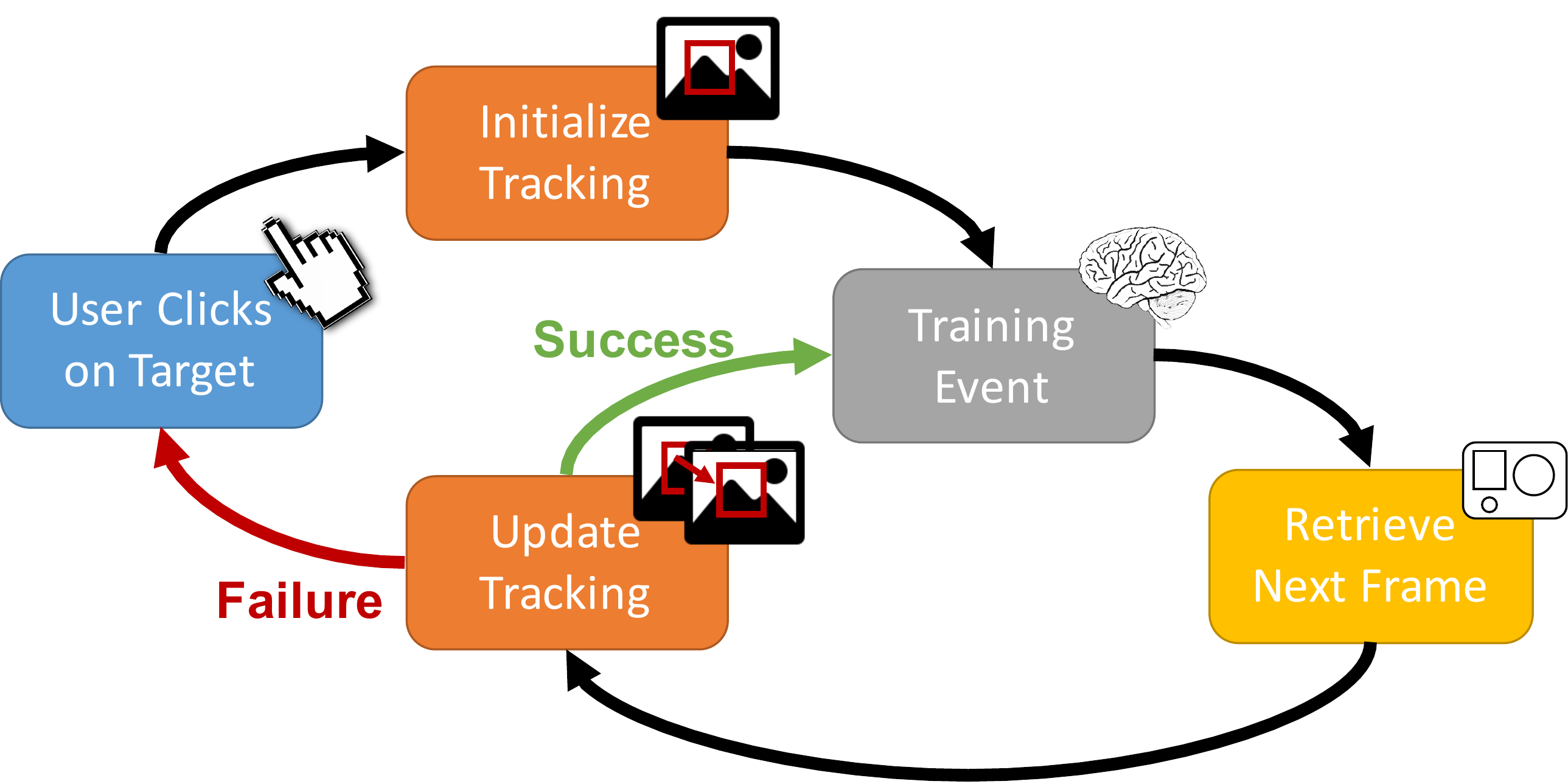}
    \caption{State diagram of object tracking-enabled online learning. Tracking is initiated by a user interaction denoting the location of the target, and continues generating new training events as long as the tracking algorithm is tracking the target.}
    \label{fig:opticalflow}
\end{figure}

To further reduce required user interactions, we exploited a property particular to ToOT sequences: the target object is unlikely to have moved far between frames. We can thus assist the user with object tracking, as long as the target item is in the scene. This tracked information can generate training events beyond what was produced by the user. Figure~\ref{fig:opticalflow} shows the method by which object tracking informs online training. When the user clicks on the target in the frame, the tracker is initialized on the target with a bounding box from segmentation. This initiates a training event. After training, a subsequent frame is retrieved from the camera, and the tracker is updated using this frame. If the tracking update succeeds, another training event is initiated with the center of the updated bounding box as the new click, and the cycle continues. If it fails, the user will have to click on the target again.

In our experiments, we use the Median Flow~\cite{Kalal2010}, because of its robust tracking failure detection.

\label{sec:experiments_weaksup}
\section{Experiments: Weakly Supervised Detection}
\label{sec:experiments_weakly}
In this section, we validate our approach to generating weak supervision through clicks. In the following Section~\ref{sec:experiments_clickbait}, we will evaluate the combined system on online training tasks.

We first compare our customized segmenter with Xu \etal\cite{Xu2016} on interactive segmentation. We fine-tune our FCN-8s segmenter on the PASCAL VOC 2012~\cite{EveringhamTheResults} training dataset. The IoU accuracy of our model using 1 to 5 clicks per object on the validation set is 0.77, where \cite{Xu2016} achieves 0.70 with a mean of 2.5 clicks. The result shows that our method achieves comparable performance without negative clicks.

\begin{figure}[h]
\centering
  \includegraphics[width=0.85\linewidth]{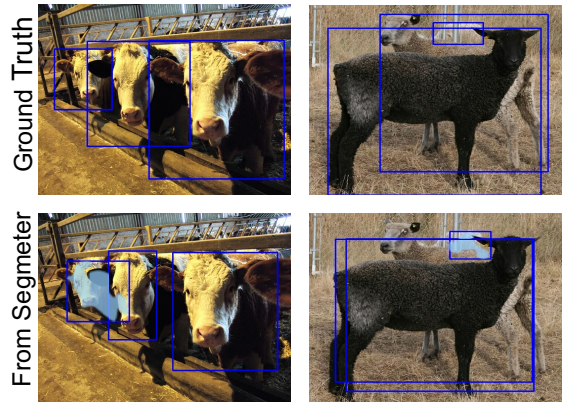}
    \caption{Sampled ground truth bounding boxes and those generated from click supervision. The resulting segmentation mask of one of the objects is overlaid on the bottom row as an example. The righthand column represents one of the most common failure modes: click ambiguity from the ground truth bounding boxes.}
    \label{fig:samples}
\end{figure}

We then use the segmenter as the only weak supervision in training a SSD300 model. To convert the fully supervised VOC dataset to a click-based one, we choose the center of each ground truth bounding box as the location of the click. For each object, we generate a click channel and input it into our segmenter (Figure~\ref{fig:segment_click}). The output mask is used to derive a bounding box---first, by removing regions not located where the click is---which is then used to supervise our SSD300 model. Although we only have simulated bounding boxes from a weak supervision of clicks, our model obtained more than half of mean average precision (mAP) of a fully supervised model and is 6\% higher than the state-of-the-art weakly supervised model that uses only class labels, as shown in Table \ref{tab:voc12}. Figure~\ref{fig:samples} shows some sample bounding boxes generated from this method. Note on the right side the most common failure mode: ambiguity of clicks. The center location of two of the sheep in the image both land on the same sheep, and cause the segmenter to produce two bounding boxes around the same object. We note that this is a limitation purely of how we generate clicks from the existing bounding boxes, and not an issue in human-driven online training; the user would \textit{not} click on overlapped portions.


\begin{table*}[ht]
	\centering
	\setlength{\tabcolsep}{3.2pt}
	\begin{tabular*}{\textwidth}{c|c|c| ccccc ccccc ccccc ccccc}
	 \scriptsize Method  & \tiny Supervision & \scriptsize mAP & \scriptsize aero & \scriptsize bike & \scriptsize bird & \scriptsize boat & \scriptsize bottle & \scriptsize bus & \scriptsize car & \scriptsize cat & \scriptsize chair & \scriptsize cow & \scriptsize table & \scriptsize dog & \scriptsize horse & \scriptsize mbike & \scriptsize person & \scriptsize plant & \scriptsize sheep & \scriptsize sofa & \scriptsize train & \scriptsize tv \\
        \hline
	\scriptsize LCL\cite{Wang2014} & \scriptsize cls & 31.6 & \textbf{48.9} & 42.3 & 26.1 & 11.3 & 11.9 & 41.3 & 40.9 &34.7 &10.8 &34.7 &18.8 &34.4 &35.4 &\textbf{52.7} &19.1 &\textbf{17.4} &35.9 &33.3 &34.8 &46.5 \\
	\scriptsize WSDDN\cite{Bilen16} & \scriptsize cls  & 34.9 & 43.6 & \textbf{50.4} & 32.2 & 26.0 & 9.8 & \textbf{58.5} & \textbf{50.4} &30.9 &7.9 &36.1 &18.2 &31.7 &41.4 &52.6 &8.8 &14.0 &37.8 &\textbf{46.9} &53.4 &47.9\\
	\scriptsize Ours &\scriptsize cls+click & \textbf{40.6} & 42.3 & 30.7 & \textbf{37.7} & \textbf{28.3} & \textbf{14.1} &52.4 &48.4 & \textbf{74.2} & \textbf{13.7} & \textbf{41.9} & \textbf{25.8} & \textbf{67.1} & \textbf{45.0} & 47.6 & \textbf{43.9} & 15.9 & \textbf{38.8} & 32.2 & \textbf{62.9} & \textbf{48.3}\\
    \hline
	\scriptsize SSD300\cite{Liu2016} &\scriptsize cls+bbox & 74.3 & 75.5 &80.2 &72.3 &66.3 &47.6 &83.0 &84.2 &86.1 &54.7 &78.3 &73.9 &84.5 &85.3 &82.6 &76.2 &48.6 &73.9 &76.0 &83.4 &74.0\\
        \noalign{\smallskip}
	\end{tabular*}
	\caption{\textbf{PASCAL VOC2007 test detection average precision (\%).} The first three are weakly supervised models. Supervision: "cls": class, "cls+click": class and bounding box derived from clicks, "cls+bbox": class and ground truth bounding box. The latter two are trained on VOC 2007 and 2012 trainval.}
    \label{tab:voc12}
\end{table*}

\section{Experiments: Online Training}
\label{sec:experiments_clickbait}
We now validate our approach for accelerating online training. In our experiments, we want to (a) show that using weakly-supervised segmentation to produce bounding boxes do not break the training process, and (b) quantify the benefit of using optical-flow assisted training for time-ordered online training of an object detector. 

\subsection{Benchmark Datasets for ToOT}

Existing datasets used in detection, such as PASCAL VOC~\cite{EveringhamTheResults} provide a rich corpus of annotated images of disparate scenes, but do not adequately represent a single training session in the field, where each image in the set directly follows the last. The KITTI suite~\cite{Geiger2012AreSuite} provides annotated video sequences, but they do not track any one particular training target, instead annotating classes of items (\eg{} cars). 

In order to validate our approach and compare online training methodologies systematically, we constructed two scenarios based on realistic use cases, each with \emph{train} and \emph{test} phases. Each phase consists of a time-ordered series of human-labeled images taken from the UAS in flight, once per second. In real-world terms, this is enough time for the UAS' embedded GPU platform to complete one training round. Each frame is annotated with three pieces of information: whether or not the target is in the frame, the coordinates of the center of the target, and the bounding box. Using these pieces of information in a frame-by-frame order, we can simulate, in a repeatable fashion, a user supervising the learning process.

\begin{figure}[h]
\centering
  \includegraphics[width=\linewidth]{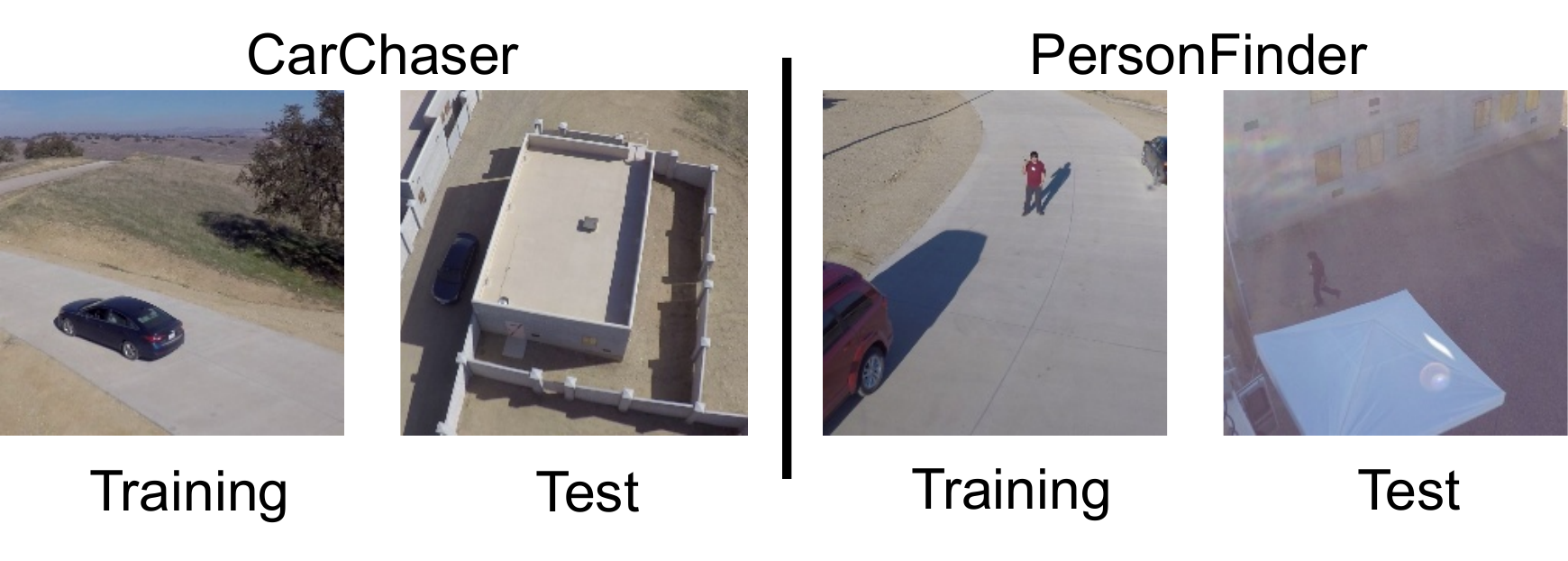}
    \caption{Sampled positive examples from CarChaser and PersonFinder training and test datasets.}
    \label{fig:datasets}
\end{figure}

We will refer to our scenarios as \emph{CarChaser} (878 train and 317 test images) and \emph{PersonFinder} (422 train and 156 test images), where the target in question is a car or a person, respectively. Images are 720x720 pixels in size. Figure~\ref{fig:datasets} shows examples from the dataset. The test scenarios are intentionally less controlled, and re-enact a sequence where a UAS is attempting to find a hidden or moving target.

\subsection{Training Strategies}

We evaluated the following training strategies on our two ToOT scenarios. 


\textbf{Online Training w/ Bounding Boxes:} In this strategy, every positive example is used in training. In other words, $u_i = 1$, where $i={1,...,n}$. Augmentation is performed as described in Section~\ref{sec:ssd}. In this case, we use manually-labeled bounding boxes as the ground truth for the SSD detector. This strategy forms a baseline for comparison of the two others. 

\textbf{Online Training w/ Click-based Supervision:} In this strategy, we performed training in the same manner as in the prior one, except the bounding boxes are generated purely from single-point clicks (Section~\ref{sec:click_seg}) near the centroid of the target. In real-world terms, this means the user only has to \textit{click} on the object during the training sequence, rather than draw a bounding box around it. 

\textbf{Online Training with Object Tracking:} In this strategy, we assisted our standardized user using Median Flow tracking, as described in Section~\ref{sec:optical_flow}. We made the following assumptions about how the user uses the tracking-assisted system:

\begin{itemize}[nosep]
\item When no tracker is initialized, and the target is in the scene, the user clicks on the target and initializes tracker. 
\item When tracker updates are successful, the user allows the system to perform training without interruption.
\item When the tracker loses the target (\ie{} the IoU of the tracked bounding box with the manually-labeled bounding box is below 0.5), the user again clicks on the target in the next frame, re-initializing the tracking. 
\end{itemize}

This means that on many occasions, $v_i$ is added to the mini-batch \textit{even if the corresponding} $u_i = 0$. This produces additional training benefit without additional user input.

\subsection{Results}

We examined two metrics---first, the maximum average precision reached by each click-supervised training strategy in comparison to using manually-labeled bounding boxes, and then the \textit{training benefit} of training with or without object tracking. All tests used as a base a 1-class SSD model pre-trained on the PASCAL VOC2007+2012 training set. Adam with $lr=0.0005, \beta_1=0.9, \beta_2=0.999, \epsilon=1*10^{-8}$ is used as the gradient descent algorithm. To mitigate the effects of stochasticity in gradient descent, all metrics were computed from a mean of 5 runs at each timestep. 

\subsubsection{Maximum Average Precision}

We used \textit{average precision}, as computed for PASCAL VOC 2007~\cite{EveringhamTheResults}, of the single class as our performance metric $P$, as seen in Equations~\ref{eqn:CTB}--\ref{eqn:ITB}.

In real-world usage, we are less concerned with finishing the entire training sequence as we are with achieving the best possible average precision, $P_{max}$, during training, where 

\begin{equation}
P_{max} = \max_{i=0}^{n} P_i
\end{equation}

In other words, $P_{max}$ is the highest achievable point when all frames to $n$ are used. We began by taking $P_{max}$ of using manually-labeled bounding boxes as ground truth, and used this as a baseline for evaluating the other measurements. We then compared this $P_{max}$ to purely click-based supervision as well as click-based supervision accelerated with object tracking.

\begin{figure} \centering
    \includegraphics[width=0.9\linewidth]{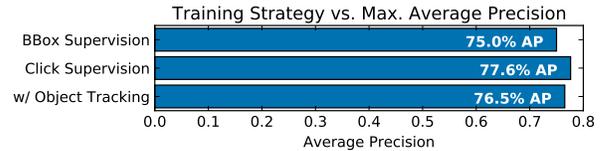}
    \caption{Comparison of $P_{max}$ between training strategies for CarChaser scenario. Using segmentation to generate bounding boxes is directly comparable to using the ground truth bounding boxes. }
    \label{fig:carchaser_acc}
\end{figure}

\begin{figure} \centering
    \includegraphics[width=0.9\linewidth]{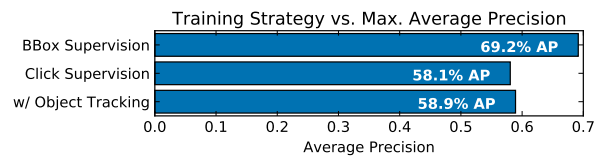}
    \caption{Comparison of $P_{max}$ between training strategies for PersonFinder scenario. Note that this scenario's test set is significantly more difficult (much smaller target) and smaller than the CarChaser scenario, making segmentation more difficult.}
    \label{fig:personfinder_acc}
\end{figure}

Figure~\ref{fig:carchaser_acc} shows this comparison for the CarChaser scenario, and Figure~\ref{fig:personfinder_acc} for the PersonFinder scenario. We see that using the click-based supervision does result in an expected but not unreasonable deviation in average precision. In PersonFinder, where the segmentation task is significantly more difficult (target is smaller, more complex), there is a noticeable degradation when using click-based supervision. On the other hand, using click-based supervision with and without object tracking produces comparable performance. This is promising as object tracking dramatically reduces the number of required clicks, as we will show. 

\begin{figure*}
\centering
\begin{minipage}[t]{.5\textwidth}
    \includegraphics[width=\textwidth]{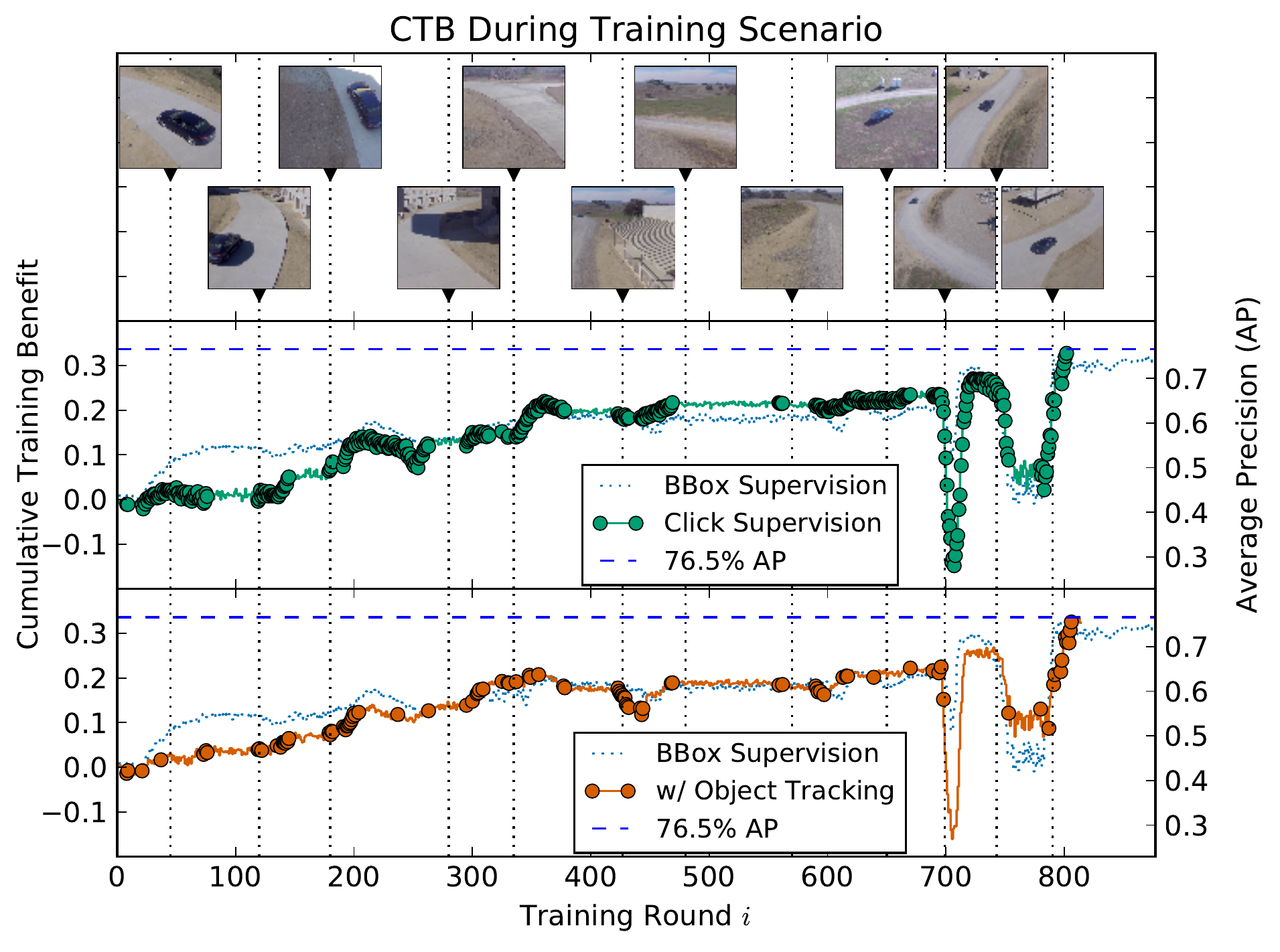}
	\caption{Cumulative training benefit for CarChaser scenario. Dots on the traces represent user interactions. Horizontal blue line represents average precision $P_f$. Training with manually-labeled bounding boxes (``BBox Supervision'') is shown as reference.}
    \label{fig:carchaser_ctb}
\end{minipage}\hfill
\begin{minipage}[t]{.46\textwidth}
    \includegraphics[width=\textwidth]{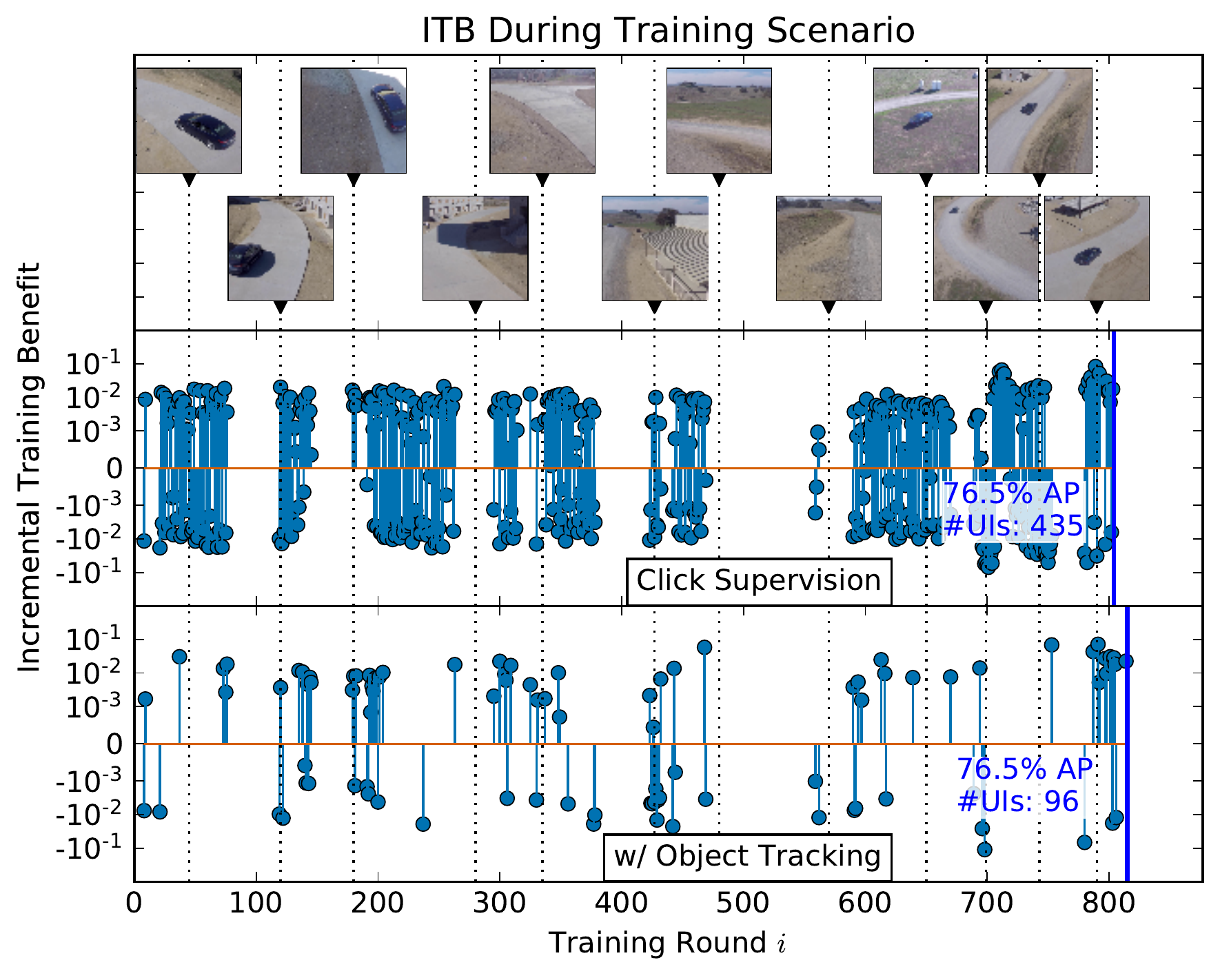}
    \caption{Incremental training benefit for CarChaser scenario. Each point represents a user interaction. Vertical line at the end represents point where accuracy reaches $P_f$. We see that the object tracking-enabled methods require far less user interactions before $P_f$ is met, although they take roughly the same length of video.}
    \label{fig:carchaser_itb}
\end{minipage}
\end{figure*}


\subsubsection{Training Benefit}

In Section~\ref{sec:background}, we defined $CTB$ and $ITB$ as metrics particular to ToOT that are a measure of the \textit{effectiveness} of each user interaction. In particular, $\overline{ITB}_{(i,j)}$ can be used to compare training strategies, given all achieve a fixed final average precision $P_f$. We assigned $P_f$ to be the minimum $P_{max}$ of the various training strategies. Runs were terminated after $P_f$ was reached. We plotted $CTB_{u_i}$ and $ITB_{u_i}$ over $v_i$, as well as computed $\overline{ITB}_{(1,f)}$, where $f$ is the step when $P_f$ is reached for that particular scenario. 

\textbf{CarChaser: }For CarChaser, the final average precision was $P_f =  76.5\%$. Training with manually-labeled bounding boxes is also shown for reference. In Figure~\ref{fig:carchaser_ctb}, we plot the CTB of both training strategies up to $P_f$. The top plot shows representative frames in the training scenario at that period of time. Each trace follows the accuracy at each training round $i$, regardless of if a user interaction happens at that step ($u_i = 1$). Each dot on the trace represents the rounds in which a user interaction did happen. 

Flat regions in the plots represent periods of no action, \ie{}  places where the target is not in the scene and the frames are not being trained on. Noise in these flat regions are a result of non-deterministic GPU execution. Note that the lines on the ``w/ object tracking'' plot closely follow those of the first plot---\textit{but the number of dots (user interactions) is much less dense.} This means that, as expected, object tracking generates user interactions that are \textit{not} triggered by the user but still effective in training the model.

We also notice dips in $CTB$; these correspond to parts of the training sequence where the training actually \textit{worsens} the accuracy. The largest trough corresponds to a part in the training sequence where the target car is very small relative to the frame. This suggests that certain properties of the frame may be predictors of their effectiveness in training the model---we leave this question for future study. 

Figure~\ref{fig:carchaser_itb} plots the incremental training benefit ($ITB$) for each user interaction in Figure~\ref{fig:carchaser_ctb}. As expected, the tracking-enabled methods produce far less dense plots as they have far fewer user interactions (Table~\ref{tab:uis}). We note that while some of the user interactions produce a very large $ITB$, the vast majority produce very small changes. To emphasize the smaller values, the y-axis is plotted on a logarithmic scale. With the object tracking-enabled methods, these low-impact training rounds can be triggered many at a time.

\begin{figure} 
\centering
    \includegraphics[width=0.9\linewidth]{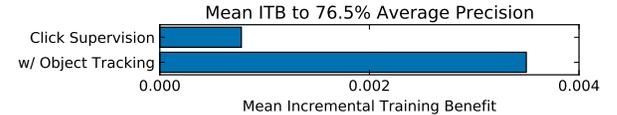}
    \caption{Mean incremental training benefit for CarChaser scenario, up to $P_f = 0.765$. Using object tracking increases $\overline{ITB}_{(1,f)}$ by over 4 times.}
    \label{fig:carchaser_meanitb}
\end{figure}

The $\overline{ITB}_{(1,f)}$ for each training technique is shown in Figure~\ref{fig:carchaser_meanitb}---the average of every point in Figure~\ref{fig:carchaser_itb}. We see that the object tracking-enabled strategy enables user interactions that are 4.50 times more effective (\ie{} produce on average 4.50 times more $ITB$ per interaction) than without.

\begin{figure*}
\centering
\begin{minipage}[t]{.5\textwidth}
    \includegraphics[width=\textwidth]{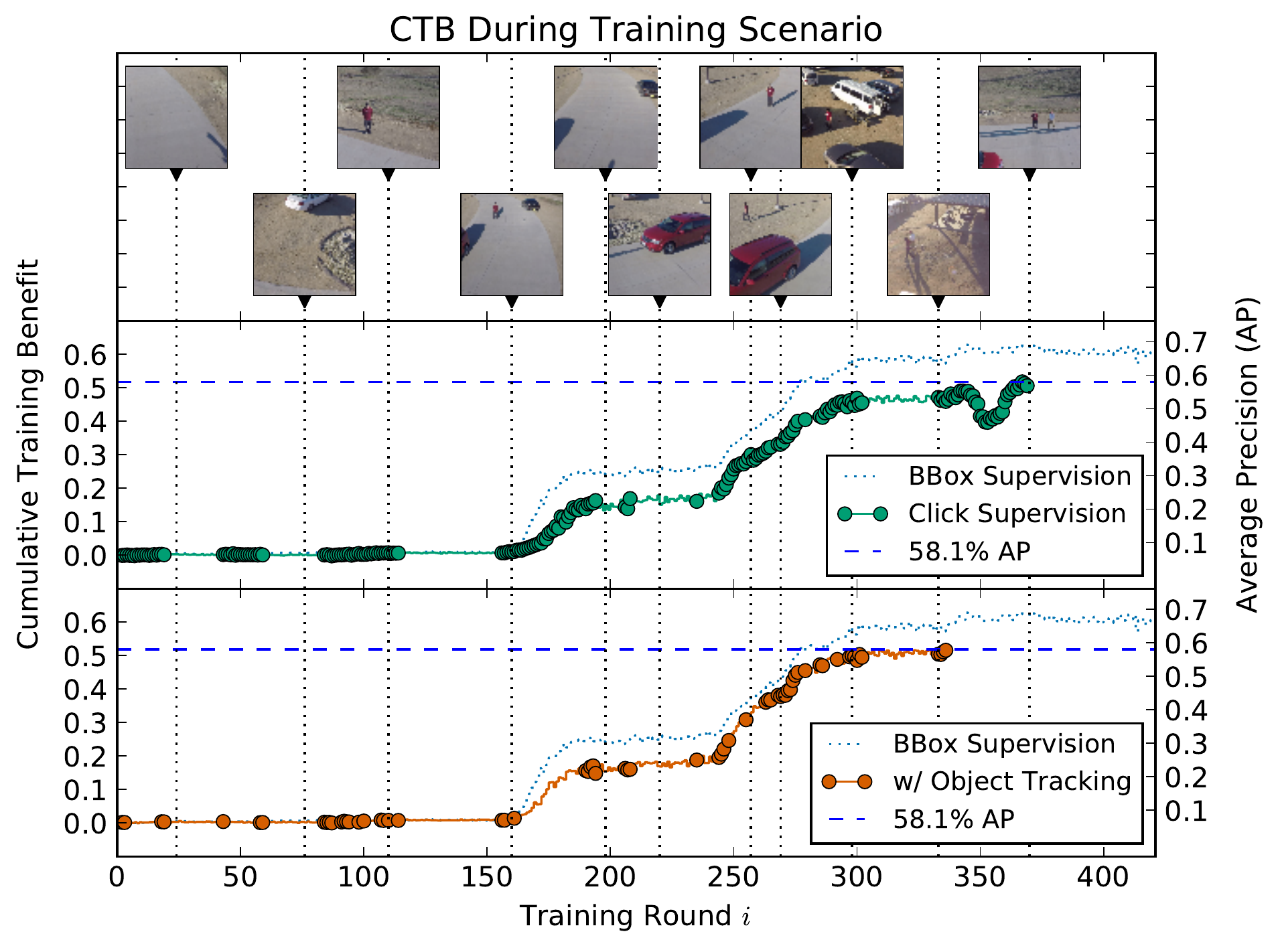}
	\caption{Cumulative training benefit for PersonFinder scenario. Dots on the traces represent user interactions. Horizontal blue line represents average precision $P_f$. We note that the number of video frames (\ie{} time) required for both are similar.}
    \label{fig:personfinder_ctb}
\end{minipage}\hfill
\begin{minipage}[t]{.46\textwidth}
    \includegraphics[width=\textwidth]{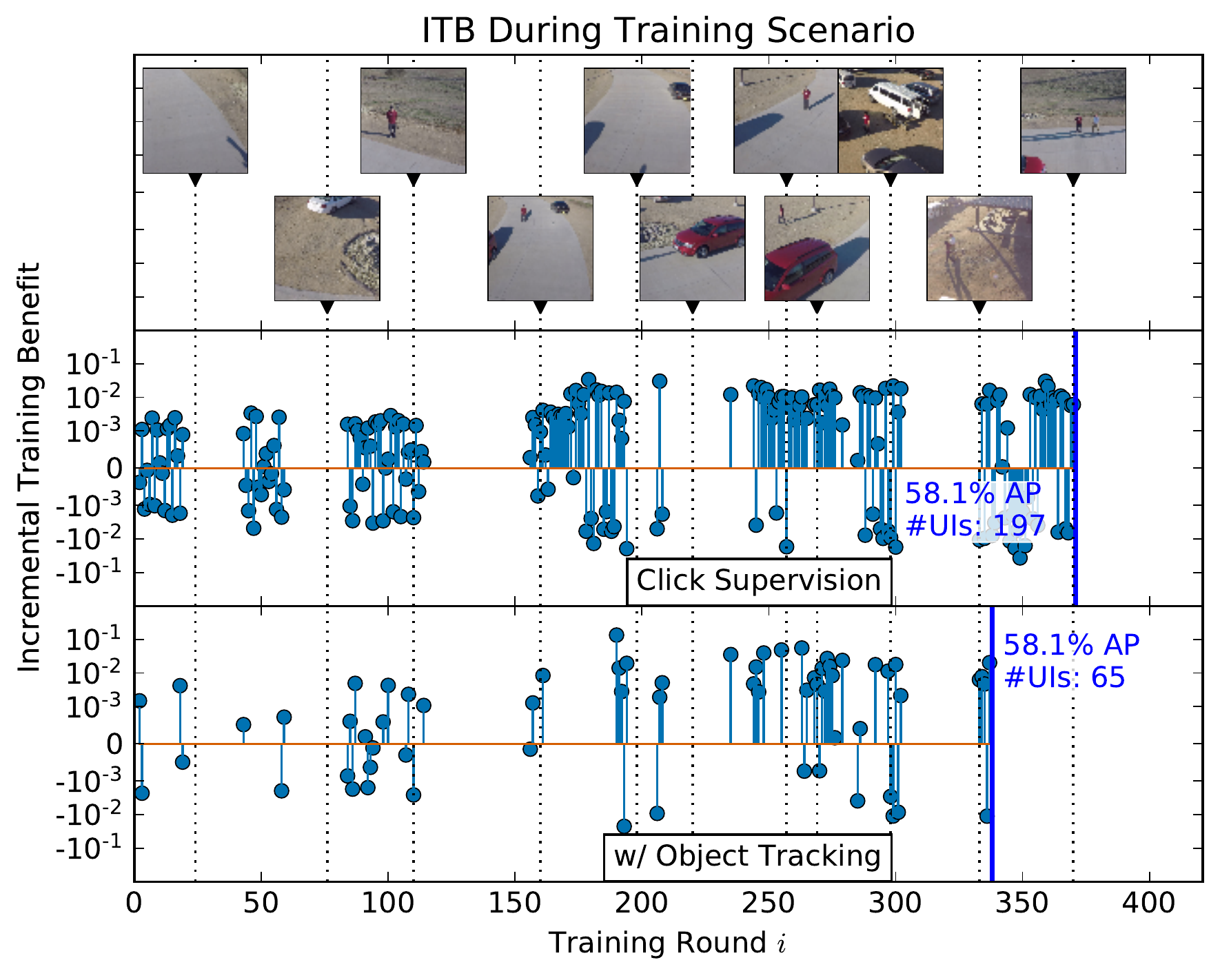}
    \caption{Incremental training benefit for PersonFinder scenario. Each point represents a user interaction. Vertical line at the end represents point where accuracy reaches $P_f$. We see that the object tracking-enabled methods require far less user interactions before $P_f$ is met, although they take roughly the same length of video.}
    \label{fig:personfinder_itb}
\end{minipage}
\end{figure*}

\begin{figure}[h]
\centering
    \includegraphics[width=0.9\linewidth]{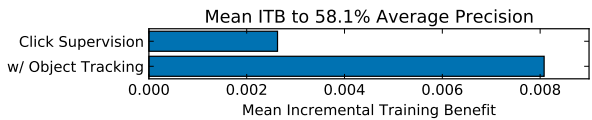}
    \caption{Mean incremental training benefit for PersonFinder scenario, up to $P_f = 58.1\%$. Object tracking increases $\overline{ITB}_{(1,f)}$ by over 3 times.}
    \label{fig:personfinder_meanitb}
\end{figure}

\textbf{PersonFinder: }We repeated the same experiments on the PersonFinder dataset. The PersonFinder test scenario contained far more clutter, and thus the initial model performed very poorly. The trained model rose from $P_0 = 6.2\%$ to $P_f = 58.1\%$.

In Figure~\ref{fig:personfinder_ctb}, enabling object tracking again reduces the density of clicks without reducing the upwards trend of performance. Interestingly, the $CTB_i$ did not climb significantly until around $i=150$, suggesting those training examples were not useful in ruling out clutter from the test set. We see the same trend reflected in the $ITB_{u_i}$ plot, with the user interactions that happened earlier having a smaller $ITB_{u_i}$ than the ones that happen later. This implies that those early training examples were less ``useful'' than those that happen later, suggesting yet again that certain frames---in this case frames with very sparse backgrounds---are less valuable than others. 

Figure~\ref{fig:personfinder_meanitb} again plots $\overline{ITB}_{(1,f)}$ for the various training strategies. From the perspective of reducing user interactions, \textit{object tracking again dramatically increased }$\overline{ITB}_{(1,f)}$, by 3.07 times. Table~\ref{tab:uis} shows both the total user interactions required to reach $P_f$, as well as the mean $ITB_{u_i}$ value for these interactions. 


\begin{table}
\begin{center}
  \begin{tabular}{  c | c | c | c | c }
     & \multicolumn{2}{c|}{\scriptsize CarChaser} & \multicolumn{2}{c}{\scriptsize PersonFinder} \\ \hline
    \scriptsize\textbf{Strategy} & $\sum\limits_{i=1}^{f} u_i $& $\overline{ITB}_{(1,f)}$ & $\sum\limits_{i=1}^{f} u_i $& $\overline{ITB}_{(1,f)}$ \\ \hline
    \scriptsize w/ Segmentation. &  435 &  \num{7.78e-4} &  197 &  \num{2.63e-3}   \\ \hline
   \scriptsize w/ Object Tracking &  96 &  \num{3.50e-3} &  65 & \num{8.08e-3}   \\ 
    \noalign{\smallskip}
  \end{tabular}
   \caption{Number of user interactions and mean $ITB$ for the different training strategies. We see that using object tracking is 3-4 times more effective than without.}
   \label{tab:uis}
  \end{center}
\end{table}


\section{Conclusion}
\label{sec:conclusion}


In this work, we recognize that training CNN object detectors in real-time via human input encompasses a unique class of problem, in which we are not only concerned with the performance of the model produced, but also the human effort, \ie{} the number of user interactions, required to train the model. We use \textit{training benefit} as a metric for measuring the impact of user interactions, and compare two training strategies in their effectiveness per user interaction. We show that we can obtain bounding box annotations from click-based supervision using interactive segmentation, eliminating the need to draw them manually during online training. Finally, by exploiting the time-ordered nature of video streams through object tracking, we can multiply mean incremental training benefit by about 3-4 times, depending on the scenario, over the one-frame-one-click approach.

We believe this work sets the stage for further investigation into real-time training on ToOT sequences. From Figures~\ref{fig:carchaser_itb} and \ref{fig:personfinder_itb}, we can observe that not all training frames produce a strong positive benefit. Further work in informing frame \textit{selection} could mitigate this problem.

{\small
\bibliographystyle{ieee}
\bibliography{Mendeley,rui}

\begin{thebibliography}{10}\itemsep=-1pt

\bibitem{BalancaSingleHttps://github.com/balancap/SSD-Tensorflow}
P.~Balanca.
\newblock {Single Shot MultiBox Detector in TensorFlow. Available at
  https://github.com/balancap/SSD-Tensorflow}.

\bibitem{Bilen16}
H.~Bilen and A.~Vedaldi.
\newblock {Weakly Supervised Deep Detection Networks}.
\newblock In {\em CVPR}, 2016.

\bibitem{Croitoru2017UnsupervisedImages}
I.~Croitoru, S.-V. Bogolin, and M.~Leordeanu.
\newblock {Unsupervised Learning From Video to Detect Foreground Objects in
  Single Images}.
\newblock In {\em ICCV}, pages 4335--4343, 2017.

\bibitem{EveringhamTheResults}
M.~Everingham, L.~van Gool, C.~K.~I. Williams, J.~Winn, and A.~Zisserman.
\newblock {The PASCAL Visual Object Classes Challenge 2012 (VOC2012) Results}.

\bibitem{Geiger2012AreSuite}
A.~Geiger, P.~Lenz, and R.~Urtasun.
\newblock {Are We Ready for Autonomous Driving? The KITTI Vision Benchmark
  Suite}.
\newblock In {\em CVPR}, 2012.

\bibitem{Guisti2016}
A.~Guisti, J.~Guzzi, D.~C. Cire{\c{s}}an, H.~Fang-Lin, J.~Rodriguez,
  F.~Fontana, M.~Faessler, C.~Forster, J.~Schmidhuber, G.~Di~Caro,
  D.~Scaramuzza, and L.~M. Gambardella.
\newblock {A Machine Learning Approach to Visual Perception of Forest Trails
  for Mobile Robots}.
\newblock {\em Robotics and Automation Letters}, 1(2):661--667, 2016.

\bibitem{Jain2016}
S.~Jain and K.~Grauman.
\newblock {Click Carving: Segmenting Objects in Video with Point Clicks}.
\newblock In {\em HCOMP}, 2016.

\bibitem{Kading2016}
C.~K{\"{a}}ding, E.~Rodner, A.~Freytag, and J.~Denzler.
\newblock {Fine-tuning Deep Neural Networks in Continuous Learning Scenarios}.
\newblock In {\em ACCV 2016 Workshop on Interpretation and Visualization of
  Deep Neural Nets}, 2016.

\bibitem{Kalal2010}
Z.~Kalal and J.~Matas.
\newblock {Forward-Backward Error: Automatic Detection of Tracking Failures}.
\newblock In {\em ICPR}, 2010.

\bibitem{Kingma2015Adam:Optimization}
D.~P. Kingma and J.~L. Ba.
\newblock {Adam: A Method for Stochastic Optimization}.
\newblock In {\em International Conference on Learning Representations (ICLR)},
  2015.

\bibitem{Kiswani2016}
A.~Kiswani, A.~Aides, and M.~Silberstein.
\newblock {Deep Learning in Aerial Systems Using Jetson}, 2016.

\bibitem{Krizhevsky2012}
A.~Krizhevsky, I.~Sutskever, and G.~E. Hinton.
\newblock {ImageNet Classification with Deep Convolutional Neural Networks}.
\newblock In {\em NIPS}, pages 1--9, 2012.

\bibitem{Liu2016}
W.~Liu, D.~Anguelov, D.~Erhan, C.~Szegedy, S.~Reed, C.~Y. Fu, and A.~C. Berg.
\newblock {SSD: Single Shot Multibox Detector}.
\newblock In {\em ECCV}, 2016.

\bibitem{Mettes2016}
Mettes et~al.
\newblock {Spot on: Action localization from pointly-supervised proposals}.
\newblock In {\em ECCV}, 2016.

\bibitem{Papadopoulos2017}
D.~P. Papadopoulos, J.~R.~R. Uijlings, F.~Keller, and V.~Ferrari.
\newblock {Training object class detectors with click supervision}.
\newblock In {\em CVPR}, 2017.

\bibitem{Szegedy2014}
C.~Szegedy, W.~Liu, Y.~Jia, P.~Sermanet, S.~Reed, D.~Anguelov, D.~Erhan,
  V.~Vanhoucke, and A.~Rabinovich.
\newblock {Going Deeper with Convolutions}.
\newblock In {\em CVPR}, pages 1--9, 2015.

\bibitem{Teng2017}
E.~Teng, J.~D. Falc{\~{a}}o, and B.~Iannucci.
\newblock {ClickBAIT: Click-based Accelerated Incremental Training of
  Convolutional Neural Networks}.
\newblock {\em CoRR}, abs/1709.0, 2017.

\bibitem{Wang2014}
C.~Wang, W.~Ren, K.~Huang, e.~D. Tan, Tieniu", T.~Pajdla, B.~Schiele, and
  T.~Tuytelaars.
\newblock {Weakly Supervised Object Localization with Latent Category
  Learning}.
\newblock In {\em ICCV}, 2014.

\bibitem{Wang2017Cost-EffectiveClassification}
K.~Wang, D.~Zhang, Y.~Li, R.~Zhang, and L.~Lin.
\newblock {Cost-Effective Active Learning for Deep Image Classification}.
\newblock {\em IEEE Transactions on Circuits and Systems for Video Technology},
  pages 1--10, 2017.

\bibitem{Xu2016}
N.~Xu, B.~Price, S.~Cohen, J.~Yang, and T.~Huang.
\newblock {Deep Interactive Object Selection}.
\newblock In {\em CVPR}, pages 373--381, 2016.

\bibitem{Yosinski2014}
J.~Yosinski, J.~Clune, Y.~Bengio, and H.~Lipson.
\newblock {How Transferable are Features in Deep Neural networks?}
\newblock In {\em NIPS}, volume~27, pages 1--9, 2014.

\end{thebibliography}
}
\end{document}